%% file: main.tex
\documentclass[journal]{IEEEtran}
\usepackage{amsmath,amsfonts}
\usepackage{algorithmic}
\usepackage{array}
\usepackage{textcomp}
\usepackage{stfloats}
\usepackage{url}
\usepackage{verbatim}
\usepackage{graphicx}
\usepackage[tight]{subfigure}
\def\BibTeX{{\rm B\kern-.05em{\sc i\kern-.025em b}\kern-.08em
    T\kern-.1667em\lower.7ex\hbox{E}\kern-.125emX}}
\usepackage{balance}

\usepackage{diagbox}
\usepackage{algorithm}
\usepackage{algorithmic}
\usepackage{mathtools}
\usepackage{bm}
\usepackage{soul,color, xcolor}
\soulregister\cite7 
\soulregister\citep7 
\soulregister\citet7 
\soulregister\ref7 
\soulregister\pageref7 
\usepackage{booktabs}
\usepackage{bbding}
\usepackage{subcaption}
\usepackage{makecell,multirow}
\usepackage{colortbl}
\definecolor{mygray}{gray}{.9}
\definecolor{mypink}{rgb}{.99,.91,.95}
\definecolor{mycyan}{cmyk}{.3,0,0,0}
\usepackage[pagebackref=true,breaklinks=true, colorlinks,bookmarks=false]{hyperref}
\usepackage[utf8]{inputenc}

\usepackage{cite}
\usepackage{pifont} 

\usepackage{amssymb}
\usepackage{flafter}
\begin{document}
\title{DTFormer: Text-Guided Semantic Alignment for RGB-D Segmentation}
\author{Ziang~Wei\,$^{2}$, Yinlong~Liu\,$^{3}$, Yan~Xia\,$^{4}$, Alois~Knoll\,$^{2}$, and Hu~Cao\,$^{1*}$
\thanks{$^*$ Hu Cao is the corresponding author of this work (hu.cao@seu.edu.cn).}
\thanks{Authors Affiliation: $^{1}$School of Automation, Southeast University, Nanjing, China, $^{2}$Chair of Robotics, Artificial Intelligence and Real-time Systems, Technical University of Munich, Germany,$^{3}$City University of Macau, China, $^{4}$ University of Science and Technology of China, China.}}
\maketitle

\begin{abstract}
    RGB-D semantic segmentation has made notable progress by fusing RGB and Depth, yet mainstream models still learn features almost exclusively from pixel-level supervision, lacking direct high-level semantic constraints. This raises a central question: can external knowledge—such as language priors—inject stronger semantic discriminability into mainstream RGB-D segmentation models? We present DTFormer, a novel tri-modal (RGB-D-Text) semantic segmentation framework. At its core is Text-guided Semantic Alignment Module (TSAM) that first encodes textual cues into a set of semantic prototypes and then explicitly aligns multi-modal RGB-D features with these prototypes at multiple encoder and decoder layers. This design imposes strong semantic regularization on representation learning, guiding the network toward more discriminative features. Extensive experiments on multiple benchmarks show that DTFormer delivers consistent gains while remaining simple and efficient. Our results demonstrate that explicit semantic alignment offers an effective and practical route to improving RGB-D semantic segmentation. The code will be released upon acceptance.
\end{abstract}

\input{Sections/1_Introduction}
\input{Sections/2_RelatedWorks}
\input{Sections/3_Method}
\input{Sections/4_Experiments}
\input{Sections/5_Conclusion}

\bibliographystyle{IEEEtran}
\bibliography{egbib}
\end{document}

%% file: Sections/1_Introduction.tex
\section{Introduction}
\label{sec:intro}

\IEEEPARstart{S}{emantic} segmentation, a core task in computer vision, assigns a semantic label to every pixel, and it underpins applications including autonomous driving \cite{caesar2020nuscenes,behley2019semantickitti,sun2020waymo}, robotics \cite{rosinol2020kimera,savva2019habitat}, augmented reality \cite{mccormac2017semanticfusion}, and medical imaging \cite{ronneberger2015unet}. With the advance of 3D sensing technologies, RGB-D data has emerged as a robust paradigm for scene understanding, particularly under occlusion and low-light conditions \cite{hu2019acnet,zhang2023cmx,dai2023scannetpp}. In parallel, large-scale vision-language models \cite{radford2021clip,li2021albef,li2022blip} have demonstrated that textual knowledge can provide class-level anchors and transferable high-level semantics for dense prediction, as evidenced in text-driven 2D segmentation \cite{li2022lseg,xu2022groupvit,rao2022denseclip}. However, a tri-modal paradigm that injects such language priors into geometrically-aware RGB-D segmentation remains surprisingly underexplored.

\begin{figure}[t!]
  \centering
  \includegraphics[width=0.8\linewidth,trim=130 140 480 60,clip]{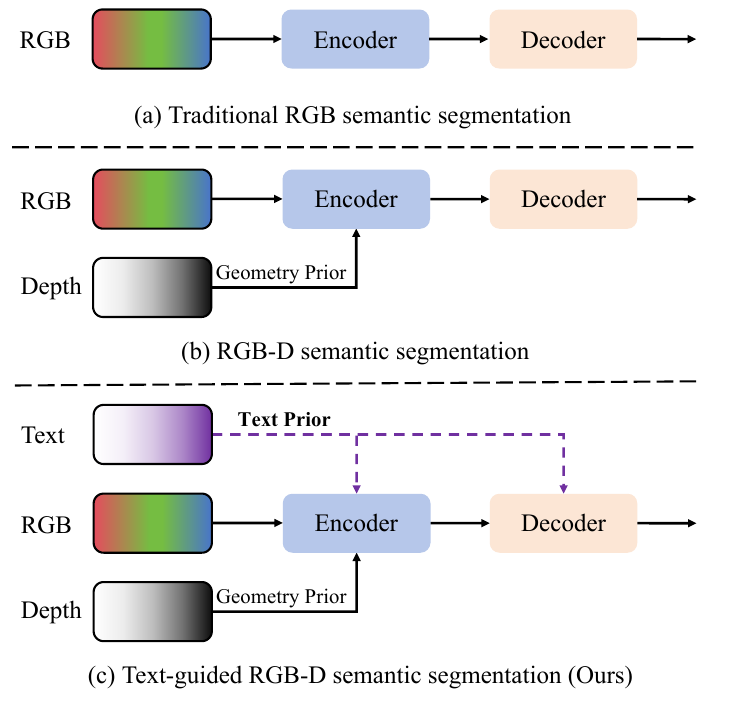}
  \caption{Comparison of semantic segmentation pipelines.
  (a) Single-stream encoder-decoder~\cite{xie2021segformer,chen2018deeplabv3plus}.
  (b) Depth as a geometric prior~\cite{YinCVPR2025DFormerv2}.
  (c) Our DTFormer injects language priors into both encoding and decoding to explicitly align RGB-D representations.}
  \label{fig:intro-pipeline}
\end{figure}

As illustrated in Fig.~\ref{fig:intro-pipeline}, the evolution of semantic segmentation can be characterized by the modalities each pipeline leverages. Conventional RGB-only methods (Fig.~\ref{fig:intro-pipeline}(a)) feed a single image to an encoder-decoder \cite{chen2018deeplabv3plus, long2015fcn}, but they remain constrained by appearance and texture cues alone, often failing under illumination changes, occlusion, or texture ambiguity. To address this, RGB-D segmentation (Fig.~\ref{fig:intro-pipeline}(b)) introduces depth as a geometric cue to alleviate such ambiguities; for instance, DFormerv2 \cite{YinCVPR2025DFormerv2} treats depth as a geometry prior that modulates self-attention, achieving a better accuracy-efficiency trade-off. However, existing RGB-D pipelines still rely mainly on pixel-level supervision and geometric constraints, largely overlooking the high-level semantic regularities available in textual form.

Recently, the success of vision-language models (VLMs) \cite{radford2021clip,li2022blip} has inspired textual priors for dense prediction. Early methods such as DenseCLIP \cite{rao2022denseclip} and LSeg \cite{li2022lseg} align visual features with fixed CLIP class-name embeddings, but applying a single dataset-level vocabulary to every image introduces semantic ambiguity from text tokens irrelevant to the actual scene. Subsequent learnable-query approaches like X-Decoder \cite{zou2023xdecoder} and SEEM \cite{zou2023seem} achieve more flexible open-vocabulary segmentation, yet, as general-purpose models, they incur additional computation and rarely model modality-specific characteristics. Moreover, most VLM-based segmentation pipelines inject language only at the decoder \cite{rao2022denseclip,li2022lseg,zou2023xdecoder,zou2023seem}, missing the opportunity to guide feature learning throughout the encoder.

\begin{figure*}[t]
  \centering
  \includegraphics[width=0.90\linewidth,trim=95 170 95 40,clip]{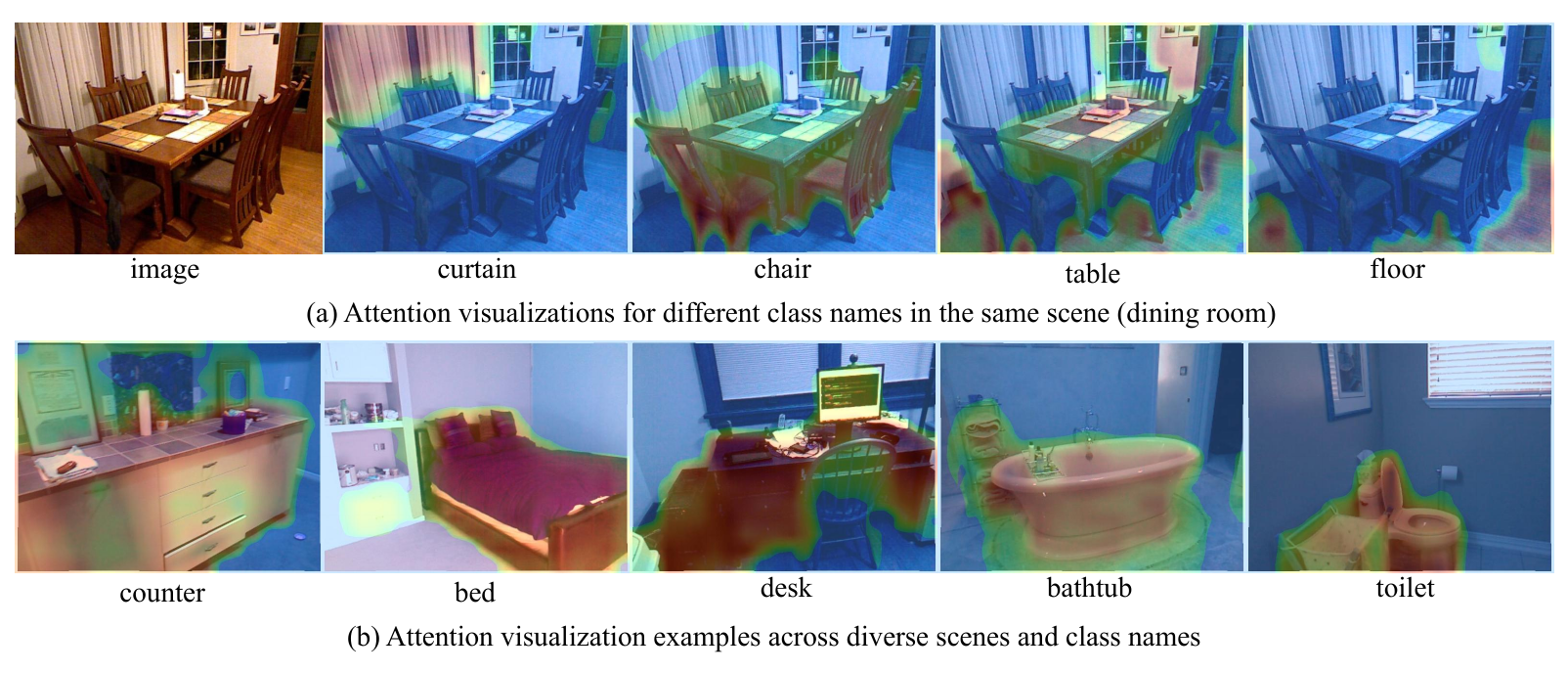}
  \caption{TSAM attention visualization.
    (a) For the same dining room scene, TSAM attends to different regions (curtain, chair, table, floor) as the target class name changes, demonstrating fine-grained semantic-visual alignment within a single image.
    (b) Across diverse indoor scenes (kitchen, bedroom, office, bathroom), TSAM consistently localizes the queried class (counter, bed, desk, bathtub, toilet), showing cross-scene generalization.
  }
  \label{fig:attn}
\end{figure*}

\begin{figure*}[!t]
  \centering
  \includegraphics[width=0.9\linewidth,trim=110 250 80 110,clip]{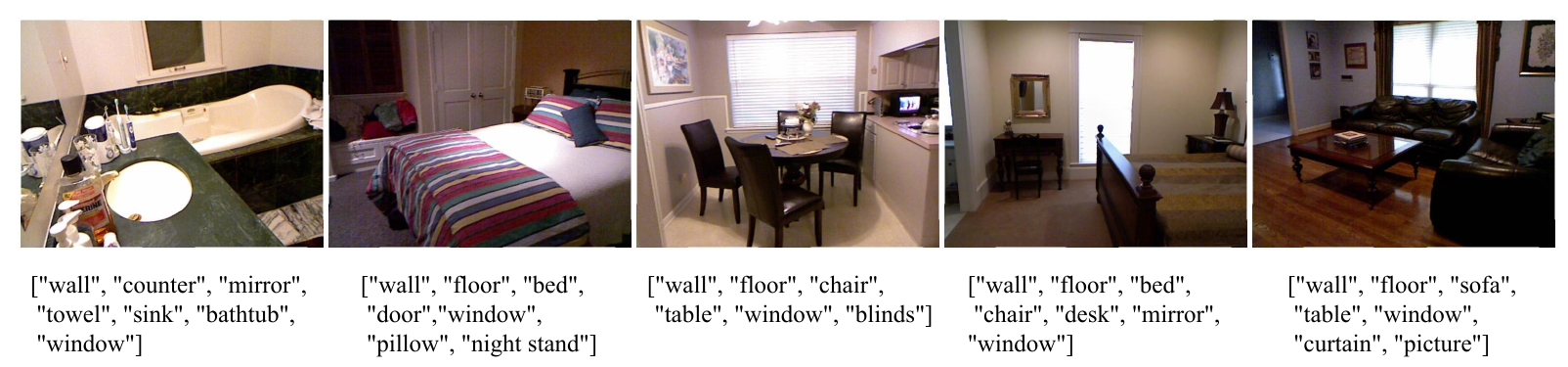}
  \caption{Examples of VLM-generated class names for diverse indoor scenes from the NYU Depth V2 dataset.
Each image is paired with its corresponding set of class names obtained through dual-VLM intersection filtering ($K=2$).}
  \label{fig:vlm_examples}
\end{figure*}

To address these limitations, we introduce the Text-guided Semantic Alignment Module (TSAM), which bridges RGB-D fused visual features and textual modalities through temperature-scaled cross-attention (TSCA) with stage-adaptive designs. As illustrated in Fig.~\ref{fig:attn}, TSAM learns to selectively attend to semantically relevant regions based on different class names, enabling fine-grained semantic-visual alignment. Building on TSAM, we construct DTFormer, a hierarchical text-guided framework for RGB-D segmentation that leverages vision-language models to generate image-specific text priors (Fig.~\ref{fig:vlm_examples}) and injects them at multiple stages throughout the network. Unlike methods that rely on fixed class names or inject language only at the decoder, DTFormer employs a dual-variant strategy: a lightweight Text-guided Semantic Alignment for Encoder (TSA-E) enables semantic feature extraction with minimal overhead, while an enriched Text-guided Semantic Alignment for Decoder (TSA-D) provides fine-grained semantic-geometric alignment with additional normalization and an MLP. Built upon the DFormerv2 \cite{YinCVPR2025DFormerv2} backbone, DTFormer effectively fuses RGB, depth geometry, and language semantics, creating a tri-modal representation for robust scene understanding.

We evaluate DTFormer on two widely-used RGB-D semantic segmentation benchmarks, NYU Depth V2~\cite{silberman2012nyuv2} and SUN-RGBD~\cite{song2015sunrgbd}. Our method achieves consistent improvements across all model scales (Tab.~\ref{tab:main_results}): DTFormer-S attains 57.8\% mIoU on NYU Depth V2 and 53.5\% on SUN-RGBD, while DTFormer-L establishes new state-of-the-art (SOTA) results with 58.6\% and 53.9\% mIoU respectively, surpassing previous best methods with competitive computational efficiency. These results validate that hierarchical injection of image-specific language priors can effectively complement geometric reasoning in RGB-D segmentation.

Our main contributions are summarized as follows:
\begin{itemize}
\item We propose the Text-guided Semantic Alignment Module (TSAM) with stage-adaptive encoder/decoder variants, injecting language priors across the network to enable hierarchical text–visual-geometry alignment.
\item We present DTFormer, a tri-modal RGB–Depth–Text framework that fuses VLM-generated, image-specific text priors with RGB-D representations via multi-stage alignment, addressing the lack of high-level semantic constraints in existing RGB-D segmentation methods.
\item Extensive experiments on NYU Depth V2 and SUN-RGBD show that DTFormer achieves new SOTA results with moderate parameter increase, validating the effectiveness of hierarchical language guidance for RGB-D semantic segmentation.
\end{itemize}

%% file: Sections/2_RelatedWorks.tex
\section{Related Work}
\label{sec:related_work}
\textbf{RGB-D Semantic Segmentation.}
RGB-only semantic segmentation often suffers from significant performance degradation in complex real-world scenarios, such as low-light conditions or cluttered environments. The integration of depth data, which provides crucial 3D geometry information, has led to the development of RGB-D semantic segmentation \cite{gupta2014learning, park2017rdfnet}. Early approaches \cite{hazirbas2017fusenet, jiang2018rednet} utilized dual-stream encoders to process RGB and depth in separate backbones, followed by various feature fusion schemes. Subsequent works introduced more sophisticated fusion mechanisms, such as designing attention-based gates \cite{hu2019acnet, chen2020sagate} or shape-aware operations \cite{cao2021shapeconv, chen2021sgnet}. With the advent of Transformers, this paradigm was extended by \cite{wang2022tokenfusion, jia2024geminifusion, zhang2023cmx}, which employ cross-modal interaction modules—e.g., cross-attention or token replacement—to exchange information between modalities. To address the high computational cost of dual-stream models, an alternative line explores efficient or more tightly coupled encoder designs \cite{du2024asymformer, yin2024dformer, seichter2022emsanet}. Most recently, DFormerv2 \cite{YinCVPR2025DFormerv2} takes this further, treating depth not as another encoded feature but as an explicit geometry prior that directly guides the self-attention weights of the RGB extractor. While these methods mark significant advancements in fusing photometric and geometric information, they rely exclusively on visual-spatial cues, overlooking the rich contextual guidance offered by language.

\textbf{Vision-Language for Dense Prediction.}
Large-scale vision-language models, particularly CLIP~\cite{radford2021clip}, have inspired text-guided dense prediction. Early works in text-guided segmentation, such as DenseCLIP \cite{rao2022denseclip}, LSeg \cite{li2022lseg}, and CLIPSeg \cite{luddecke2022clipseg}, align pixel-level features with text embeddings and are often instantiated with dataset-level vocabularies. While effective, this approach introduces semantic ambiguity, as the model must disambiguate against irrelevant text tokens.
This limitation motivated efforts to replace fixed, dataset-level vocabularies with more flexible text inputs. One popular approach \cite{zou2023xdecoder, zou2023seem, xu2023odise, li2024omgseg} utilizes learnable text queries to segment arbitrary concepts. A parallel line of work \cite{zhong2022regionclip, li2022glip, Kamath2021MDETR} focuses on text-to-box grounding to achieve region-level segmentation. Despite their power, these methods share two fundamental limitations in the context of our research: (1) they are designed as general-purpose 2D models, overlooking modality-specific characteristics such as depth geometry, and (2) a common choice is to inject language priors primarily at the decoder stage, which may miss the opportunity to guide early-stage feature extraction. Unlike these approaches, DTFormer generates image-specific class names via VLMs and injects text priors hierarchically in both encoder and decoder stages, introducing text guidance atop geometry-aware RGB-D backbones.

%% file: Sections/3_Method.tex
\section{Method}

\subsection{Image-Specific Text Generation}
\label{sec:text_generation}
\begin{figure*}[t]
  \centering
  \includegraphics[width=0.90\textwidth,trim=50 0 115 0,clip]{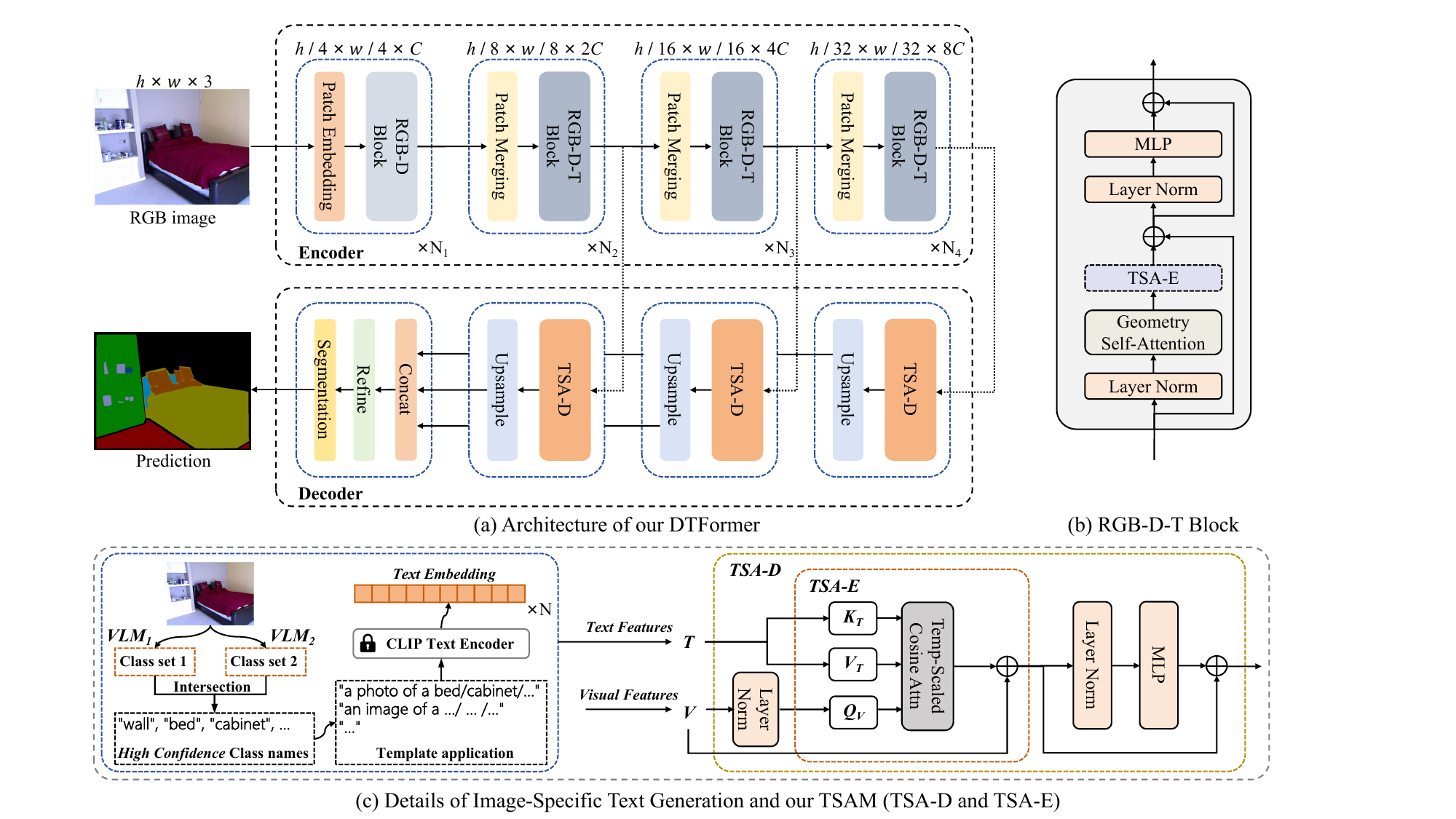}
  \caption{
    Overall architecture of our DTFormer.
    (a) Four-stage encoder and three-stage decoder. The encoder extracts multi-scale features, while the decoder progressively upsamples and refines segmentation predictions.
    (b) RGB-D-T block in the encoder, where TSA-E is applied after Geometry Self-Attention (GSA).
    (c) Details of image-specific text generation and TSAM variants. Left: Two different VLMs query the input image with the same prompt, and intersection filtering retains only class names predicted by both models to reduce hallucination, yielding approximately 4--8 high-confidence class names per image. These class names are wrapped into templates and encoded by frozen CLIP text encoder. Right: TSA-D (decoder) employs enriched Transformer blocks with Pre-LayerNorm, temperature-scaled cosine attention, and MLP for fine-grained refinement, while TSA-E (encoder) uses lightweight temperature-scaled cosine attention only for efficient semantic injection. Both variants use learnable gating for gradual semantic alignment.
  }
  \label{fig:architecture}
\end{figure*}

Unlike methods that rely on fixed class names or learnable generic queries, our approach leverages vision—language models (VLMs) to generate \emph{image-specific} text priors tailored to each input. Fig.~\ref{fig:architecture}(c) illustrates this offline text generation pipeline, where VLM predictions are encoded into CLIP \cite{radford2021clip} text embeddings for downstream semantic alignment. Given an RGB image $\mathcal{I}_{\text{RGB}}$ (depth is utilized only in the downstream segmentation network), we query $K$ different VLMs using the same prompt $p$, with output vocabulary constrained to the dataset's canonical class set $\mathcal{A}$:
\begin{equation}
L_i = \mathrm{VLM}_{i,\mathcal{A}}(\mathcal{I}_{\text{RGB}}; p) \subseteq \mathcal{A}, \quad i=1,\dots,K,
\end{equation}
where $\mathrm{VLM}_{i,\mathcal{A}}(\cdot;\cdot)$ denotes the $i$-th VLM's output restricted to vocabulary $\mathcal{A}$.
To ensure consistency and reduce hallucination, we apply an intersection-based filter across all VLMs:
\begin{equation}
L = \bigcap_{i=1}^{K} L_i .
\end{equation}
This simple yet effective mechanism retains only the class names predicted by all $K$ VLMs, thereby improving precision and reducing model-specific biases.

\begin{table*}[!t]
\centering
\caption{Comparison with SOTA methods on NYU Depth V2 and SUN-RGBD. Methods are grouped by scale (small, base, large) for fair comparison.}
\label{tab:main_results}
\small
\setlength{\tabcolsep}{9pt}
\renewcommand{\arraystretch}{1.08}
\begin{tabular}{lcccccccc}
\toprule
\textbf{Model} & \textbf{Backbone} & \textbf{Params} &
\multicolumn{3}{c}{\textbf{NYU Depth V2}} &
\multicolumn{3}{c}{\textbf{SUN-RGBD}} \\
\cmidrule(lr){4-6}\cmidrule(lr){7-9}
 &  &  & \textbf{Input} & \textbf{FLOPs} & \textbf{mIoU} & \textbf{Input} & \textbf{FLOPs} & \textbf{mIoU} \\
\midrule
TokenFusion$_{22}$~\cite{wang2022tokenfusion} & MiT-B2 & 26.0M & 480$\times$640 & 55.2G & 53.3 & 530$\times$730 & 71.1G & 50.3 \\
Omnivore$_{22}$~\cite{girdhar2022omnivore} & Swin-Tiny & 29.1M & 480$\times$640 & 32.7G & 49.7 & 530$\times$730 & --- & --- \\
DFormer$_{24}$~\cite{yin2024dformer} & DFormer-Tiny & 6.0M & 480$\times$640 & 11.7G & 51.8 & 530$\times$730 & 15.0G & 48.8 \\
DFormer$_{24}$~\cite{yin2024dformer} & DFormer-Small & 18.7M & 480$\times$640 & 25.6G & 53.6 & 530$\times$730 & 33.0G & 50.0 \\
DFormer$_{24}$~\cite{yin2024dformer} & DFormer-Base & 29.5M & 480$\times$640 & 41.9G & 55.6 & 530$\times$730 & 54.0G & 51.2 \\
AsymFormer$_{24}$~\cite{du2024asymformer} & MiT-B0+CNX-T & 33.0M & 480$\times$640 & 39.4G & 55.3 & 530$\times$730 & 52.6G & 49.1 \\
DFormerv2-S$_{25}$~\cite{YinCVPR2025DFormerv2} & DFormerv2-Small & 26.7M & 480$\times$640 & 33.9G & 56.0 & 530$\times$730 & 43.7G & 51.5 \\
\rowcolor{gray!15}
\textbf{DTFormer-S (Ours)} & DTFormer-Small & 35.5M & 480$\times$640 & 40.2G & \textbf{57.8} & 530$\times$730 & 51.9G & \textbf{53.5} \\
\midrule
SGNet$_{21}$~\cite{chen2021sgnet} & ResNet-101 & 64.7M & 480$\times$640 & 108.5G & 51.1 & 530$\times$730 & 151.5G & 48.6 \\
ShapeConv$_{21}$~\cite{cao2021shapeconv} & ResNeXt-101 & 86.8M & 480$\times$640 & 124.6G & 51.3 & 530$\times$730 & 161.8G & 48.6 \\
FRNet$_{22}$~\cite{zhou2022frnet} & ResNet-34 & 85.5M & 480$\times$640 & 115.6G & 53.6 & 530$\times$730 & 150.0G & 51.8 \\
EMSANet$_{22}$~\cite{seichter2022emsanet} & ResNet-34 & 46.9M & 480$\times$640 & 45.4G & 51.0 & 530$\times$730 & 58.6G & 48.4 \\
TokenFusion$_{22}$~\cite{wang2022tokenfusion} & MiT-B3 & 45.9M & 480$\times$640 & 94.4G & 54.2 & 530$\times$730 & 122.1G & 51.4 \\
Omnivore$_{22}$~\cite{girdhar2022omnivore} & Swin-Small & 51.3M & 480$\times$640 & 59.8G & 52.7 & 530$\times$730 & --- & --- \\
CMX$_{23}$~\cite{zhang2023cmx} & MiT-B2 & 66.6M & 480$\times$640 & 67.6G & 54.4 & 530$\times$730 & 86.3G & 49.7 \\
DFormer$_{24}$~\cite{yin2024dformer} & DFormer-Large & 39.0M & 480$\times$640 & 65.7G & 57.2 & 530$\times$730 & 84.5G & 52.5 \\
GeminiFusion$_{24}$~\cite{jia2024geminifusion} & MiT-B3 & 75.8M & 480$\times$640 & 138.2G & 56.8 & 530$\times$730 & 179.0G & 52.7 \\
DFormerv2-B$_{25}$~\cite{YinCVPR2025DFormerv2} & DFormerv2-Base & 53.9M & 480$\times$640 & 67.2G & 57.7 & 530$\times$730 & 86.9G & 52.8 \\
\rowcolor{gray!15}
\textbf{DTFormer-B (Ours)} & DTFormer-Base & 67.7M & 480$\times$640 & 79.6G & \textbf{58.3} & 530$\times$730 & 103.0G & \textbf{53.8} \\
\midrule
SA-Gate$_{20}$~\cite{chen2020sagate} & ResNet-101 & 110.9M & 480$\times$640 & 193.7G & 52.4 & 530$\times$730 & 250.1G & 49.4 \\
CEN$_{20}$~\cite{cen2020neurips} & ResNet-101 & 118.2M & 480$\times$640 & 618.7G & 51.7 & 530$\times$730 & 790.3G & 50.2 \\
CEN$_{20}$~\cite{cen2020neurips} & ResNet-152 & 133.9M & 480$\times$640 & 664.4G & 52.5 & 530$\times$730 & 849.7G & 51.1 \\
PGDENet$_{22}$~\cite{zhou2023pgdenet} & ResNet-34 & 100.7M & 480$\times$640 & 178.8G & 53.7 & 530$\times$730 & 229.1G & 51.0 \\
MultiMAE$_{22}$~\cite{bachmann2022multimae} & ViT-Base & 95.2M & 480$\times$640 & 267.9G & 56.0 & 640$\times$640 & 267.9G & 51.1 \\
Omnivore$_{22}$~\cite{girdhar2022omnivore} & Swin-Base & 95.7M & 480$\times$640 & 109.3G & 54.0 & 530$\times$730 & --- & --- \\
CMX$_{23}$~\cite{zhang2023cmx} & MiT-B4 & 139.9M & 480$\times$640 & 134.3G & 56.3 & 530$\times$730 & 173.8G & 52.1 \\
CMX$_{23}$~\cite{zhang2023cmx} & MiT-B5 & 181.1M & 480$\times$640 & 167.8G & 56.9 & 530$\times$730 & 217.6G & 52.4 \\
CMNeXt$_{23}$~\cite{zhang2023cmnext} & MiT-B4 & 119.6M & 480$\times$640 & 131.9G & 56.9 & 530$\times$730 & 170.3G & 51.9 \\
GeminiFusion$_{24}$~\cite{jia2024geminifusion} & MiT-B5 & 137.2M & 480$\times$640 & 256.1G & 57.7 & 530$\times$730 & 332.4G & 53.3 \\
DFormerv2-L$_{25}$~\cite{YinCVPR2025DFormerv2} & DFormerv2-Large & 95.5M & 480$\times$640 & 124.1G & 58.4 & 530$\times$730 & 160.5G & 53.3 \\
\rowcolor{gray!15}
\textbf{DTFormer-L (Ours)} & DTFormer-Large & 114.0M & 480$\times$640 & 161.1G & \textbf{58.6} & 530$\times$730 & 208.2G & \textbf{53.9} \\
\bottomrule
\end{tabular}

\end{table*}

\paragraph*{Exploiting scene sparsity}
Indoor scenes typically contain only a small subset of the full class set. To exploit this natural sparsity, we impose a weak cardinality constraint by instructing each VLM to output at most 8 class names ($m_{\max}=8$). With two VLMs ($K=2$), the intersection $L$ typically yields a stable set of 4--8 class names per image, balancing precision and recall.

\paragraph*{VLM implementation}
We employ two high-performance vision-language models from the InternVL~\cite{chen2024internvl} and Qwen-VL~\cite{bai2023qwenvl} families as our dual VLMs ($K=2$). Both models query each image with a fixed prompt template that instructs them to output up to 8 class names from the dataset vocabulary, prioritizing large and obvious regions. The prompt explicitly constrains outputs to match exact spellings from the provided vocabulary to minimize hallucination. Intersection filtering across both VLM outputs ensures only consistently predicted class names are retained, reducing model-specific biases while maintaining sufficient semantic coverage.

\paragraph*{VLM query prompt}
To ensure reliable and vocabulary-constrained outputs, we design a structured prompt that instructs VLMs to generate class names based on image content while prioritizing large and obvious regions.
The core prompt structure (simplified for clarity) is:

\noindent\begin{minipage}[b]{\linewidth}
\small
\begin{verbatim}
Input: One RGB image + fixed vocabulary

Goal: Return ONLY a JSON array of strings
(no code block, no prose) with UP TO 8
labels that correspond to the LARGEST and
MOST OBVIOUS regions in the image.

Selection rules:
- Include only labels that are obvious
  and match with high confidence;
- Use EXACT spelling from the vocabulary;
- DO NOT invent new labels;
- ...

Vocabulary: [wall, floor, cabinet, ...]
Output format example:
["wall", "floor", "table", ...]
\end{verbatim}
\end{minipage}\par\medskip
This simplified representation captures the essential prompt logic. The full implementation includes additional constraints to prevent hallucination and ensure deterministic JSON parsing.

\paragraph*{VLM-generated class names}
Fig.~\ref{fig:vlm_examples} presents representative examples of VLM-generated class names across diverse indoor scenes from the NYU Depth V2 dataset.
Our pipeline queries two VLMs independently with the same prompt, and intersection filtering ($K=2$) retains only class names predicted by both models.
This consensus-based approach typically yields 4--8 high-confidence class names per image, substantially reducing semantic ambiguity compared to fixed 40-class vocabulary while maintaining sufficient coverage for effective text guidance.

\paragraph*{Encoding into CLIP text space}
As shown in the left panel of Fig.~\ref{fig:architecture}(c), the filtered class names are encoded via the CLIP text encoder. Following DenseCLIP~\cite{rao2022denseclip}, we apply a template ensemble strategy: each class name $l_j$ is wrapped into multiple contextualized prompts (e.g., ``\texttt{a photo of a ...}'' and ``\texttt{an image of a ...}''), encoded via the CLIP text encoder $\mathcal{E}_{\text{text}}$, and the resulting embeddings are averaged to produce $\mathbf{t}_j$. The full text embedding matrix is:
\begin{equation}
\mathbf{T} = [\,\mathbf{t}_1,\ldots,\mathbf{t}_N\,] \in \mathbb{R}^{N \times C_t},
\end{equation}
where $C_t=512$ is the CLIP text embedding dimension and $N=|L|$ is the number of retrieved class names. The resulting matrix $\mathbf{T}$ serves as semantic guidance throughout the network, helping suppress irrelevant visual tokens and stabilizing cross-modal alignment. Although CLIP text embeddings are trained to align with CLIP image features, DTFormer does not require RGB-D segmentation features to lie in CLIP image space. In TSAM, visual features are projected into queries, while text embeddings are projected into keys and values; cross-modal attention is then computed in this attention space and optimized by the segmentation objective.

\subsection{Text-guided Semantic Alignment Module}
The core of our framework is the Text-guided Semantic Alignment Module (TSAM), which bridges visual and textual modalities through cross-modal attention. Fig.~\ref{fig:architecture}(c) illustrates the dual-variant design of TSAM. A key design challenge is balancing expressiveness and efficiency: encoder layers require early semantic injection to guide hierarchical feature learning, while decoder layers demand richer transformations.

To address this, we propose a stage-adaptive dual-variant design. Both TSA-E and TSA-D employ temperature-scaled cosine attention for stable cross-modal alignment, but differ in architectural complexity. TSA-E adopts a lightweight design with only cross-attention and residual integration, injecting semantic priors efficiently during feature extraction. TSA-D extends this with Pre-LayerNorm and MLP layers, enabling richer feature transformations for robust semantic-geometric alignment. This asymmetric design allows our model to leverage language guidance throughout the hierarchy while maintaining computational efficiency critical for RGB-D segmentation.

\paragraph*{Core attention mechanism}
Given RGB-D fused visual features $\mathbf{V}\in\mathbb{R}^{B\times H\times W\times C_v}$ (where $B$ is the batch size) and text embeddings $\mathbf{T}\in\mathbb{R}^{B\times N\times C_t}$ where $N$ denotes image-specific class names (Sec.~\ref{sec:text_generation}), we reshape $\mathbf{V}$ into $B\times HW\times C_v$ to treat each spatial location as a query. Linear projections produce queries $\mathbf{Q}_V\in\mathbb{R}^{B\times HW\times d}$, keys $\mathbf{K}_T\in\mathbb{R}^{B\times N\times d}$, and values $\mathbf{V}_T\in\mathbb{R}^{B\times N\times d}$, where $d$ is the head dimension.
We adopt temperature-scaled cosine attention for stable cross-modal alignment. Queries and keys are L2-normalized before computing similarity:
\begin{equation}
\mathbf{A} = \text{softmax}\left(\frac{\mathbf{Q}_V'\mathbf{K}_T'^T}{\tau}\right),
\end{equation}
where $\mathbf{Q}_V' = \mathbf{Q}_V/\|\mathbf{Q}_V\|_2$, $\mathbf{K}_T' = \mathbf{K}_T/\|\mathbf{K}_T\|_2$, and $\tau$ is a learnable temperature parameter. Cosine similarity stabilizes gradients by removing magnitude sensitivity, while learnable temperature enables adaptive attention sharpness. As visualized in Fig.~\ref{fig:attn}, this attention mechanism enables TSAM to selectively focus on semantically relevant regions for each class name across diverse indoor scenarios. The aligned features are computed via weighted aggregation:
\begin{equation}
\mathbf{F} = \mathbf{A}\mathbf{V}_T,
\end{equation}
where each head produces semantically aligned features. Multi-head outputs are concatenated and projected back to $C_v$, yielding $\mathbf{F}\in\mathbb{R}^{B\times HW\times C_v}$ for residual integration.

\paragraph*{TSA-E for encoder stages}
As shown in Fig.~\ref{fig:architecture}(b), TSA-E is integrated into the RGB-D-T block after geometry self-attention. After temperature-scaled cross-attention (TSCA), the aligned features are directly integrated through residual gating:
\begin{equation}
\mathbf{V}_{\text{out}} = \mathbf{V} + \beta_e \cdot \mathbf{F},
\end{equation}
where $\beta_e$ is a learnable scalar for controlled semantic injection. This lightweight design—using only cross-attention without layer normalization or MLP—preserves computational efficiency while guiding feature learning alongside geometry self-attention (GSA). TSA-E is applied at encoder stages 2, 3, and 4, skipping stage 1 which consumes more computational resources but brings limited performance improvement due to too much low-level information and higher resolution.

\paragraph*{TSA-D for decoder stages}
TSA-D employs an enriched Transformer block architecture for robust semantic-geometric alignment, as illustrated in Fig.~\ref{fig:architecture}(c). Following standard practice, we adopt Pre-LayerNorm configuration with residual connections:
\begin{equation}
\mathbf{V}' = \mathbf{V} + \beta_d \cdot \text{TSCA}(\text{LN}(\mathbf{V}), \mathbf{T}),
\end{equation}
\begin{equation}
\mathbf{V}_{\text{out}} = \mathbf{V}' + \text{MLP}(\text{LN}(\mathbf{V}')),
\end{equation}
where $\text{LN}(\cdot)$ denotes layer normalization, $\text{TSCA}(\cdot)$ is the temperature-scaled cosine attention, and $\text{MLP}(\cdot)$ is a two-layer feed-forward network with expansion ratio 4 and GELU activation. $\beta_d$ is a learnable scalar for adaptive feature integration. Layer normalization stabilizes feature distributions before attention and MLP operations, enabling more expressive transformations for robust semantic-geometric alignment. TSA-D is applied at all three decoder levels to refine features through text-visual interactions.

\subsection{Network Architecture}
Fig.~\ref{fig:architecture}(a) presents the overall architecture of DTFormer, which adopts a hierarchical encoder-decoder structure built upon the DFormerv2 backbone~\cite{YinCVPR2025DFormerv2}. The encoder extracts multi-scale features at four stages with resolutions $\{h/4, h/8, h/16, h/32\}$. Stage 1 contains RGB-D blocks with geometry self-attention, while stages 2--4 extend to RGB-D-T blocks (Fig.~\ref{fig:architecture}(b)) by integrating TSA-E for text-guided feature extraction. The decoder applies TSA-D at each pyramid level before feature fusion. After TSA-D, features are upsampled to a common resolution and fused through a Ham-style decoder head~\cite{geng2021hamnet}. A final segmentation head produces per-pixel predictions.

%% file: Sections/4_Experiments.tex
\section{Experiments}
\label{sec:experiments}

\subsection{Experimental Setup}
\label{sec:experimental_setup}

\begin{table}[!htbp]
\centering
\caption{RGB-D benchmark statistics.}
\label{tab:datasets}
\small
\renewcommand{\arraystretch}{1.125}
\begin{tabular}{l|cc}
\toprule
Attribute & NYU Depth V2 & SUN-RGBD \\
\midrule
Total images & 1,449 & 10,335 \\
Training images & 795 & 5,285 \\
Testing images & 654 & 5,050 \\
Categories & 40 & 37 \\
Resolution & 480$\times$640 & 530$\times$730 \\
Metric & mIoU & mIoU \\
\bottomrule
\end{tabular}

\end{table}

\paragraph*{Datasets}
We evaluate DTFormer on two widely-used RGB-D semantic segmentation benchmarks: NYU Depth V2~\cite{silberman2012nyuv2} and SUN-RGBD~\cite{song2015sunrgbd}, as summarized in Tab.~\ref{tab:datasets}.

\paragraph*{Implementation details}
We build DTFormer upon DFormerv2~\cite{YinCVPR2025DFormerv2} with ImageNet-1K pre-trained weights.
The CLIP ViT-B/16~\cite{radford2021clip} text encoder is frozen to extract 512-dimensional embeddings from VLM-generated class names. We use InternVL3-38B~\cite{chen2024internvl} and Qwen3-VL-30B~\cite{bai2023qwenvl} as our two VLMs ($K=2$) with intersection filtering, yielding approximately 4--8 class names per image. Text embeddings are pre-encoded and cached offline. Training employs AdamW~\cite{loshchilov2019adamw} optimizer with standard augmentations (random horizontal flipping and random scaling in [0.5, 2.0]). All experiments are conducted on 2 NVIDIA H100 GPUs. Following standard practice~\cite{YinCVPR2025DFormerv2}, we report single-scale inference with horizontal flipping on NYU Depth V2, and multi-scale inference (scales=[0.5, 0.75, 1.0, 1.25, 1.5]) with horizontal flipping on SUN-RGBD.

\begin{table}[!htbp]
\centering
\caption{Text guidance configuration in DTFormer.}
\label{tab:text_config}
\small
\renewcommand{\arraystretch}{1.125}
\begin{tabular}{lc}
\toprule
Configuration & Value \\
\midrule
Text encoder & CLIP ViT-B/16 \\
Embedding dimension & 512 \\
Encoder stages (TSA-E) & 2, 3, 4 \\
Decoder levels (TSA-D) & All 3 levels \\
VLM number $K$ & 2 \\
Typical class names per image & 4--8 \\
\bottomrule
\end{tabular}

\end{table}

\paragraph*{Text guidance configuration}
Tab.~\ref{tab:text_config} summarizes the key hyperparameters for text guidance in DTFormer. The encoder employs TSA-E at stages 2--4, while the decoder applies TSA-D at all three upsampling levels.

\begin{table}[!t]
\centering
\caption{Training configurations for DTFormer variants on NYU Depth V2 and SUN-RGBD datasets. Multiple batch sizes and input sizes are for NYU Depth V2 and SUN-RGBD datasets respectively.}
\label{tab:training}
\small
\renewcommand{\arraystretch}{1.125}
\setlength{\tabcolsep}{2pt}
\begin{tabular}{@{}lccc@{}}
\toprule
Configuration & DTFormer-S & DTFormer-B & DTFormer-L \\
\midrule
Input size & \multicolumn{3}{c}{480$\times$640 / 480$^2$} \\
Optimizer & \multicolumn{3}{c}{AdamW} \\
Base learning rate & \multicolumn{3}{c}{$6\times10^{-5}$} \\
Weight decay & \multicolumn{3}{c}{0.01} \\
Batch size & 16 / 16 & 8 / 16 & 16 / 16 \\
Epochs & \multicolumn{3}{c}{500 / 300} \\
Optimizer momentum & \multicolumn{3}{c}{$\beta_1, \beta_2 = 0.9, 0.999$} \\
Learning rate schedule & \multicolumn{3}{c}{Polynomial decay} \\
Warmup epochs & \multicolumn{3}{c}{10} \\
Warmup schedule & \multicolumn{3}{c}{Linear} \\
Layer-wise lr decay & \multicolumn{3}{c}{None} \\
Auxiliary head & \multicolumn{3}{c}{None} \\
\bottomrule
\end{tabular}

\end{table}

\paragraph*{Training configurations}
Tab.~\ref{tab:training} details the training hyperparameters for all DTFormer variants on both benchmarks.
Multiple batch sizes and input sizes are for NYU Depth V2 and SUN-RGBD datasets respectively.

\begin{table}[!t]
\centering
\caption{Detailed configurations of DTFormer variants. $C_i$ represents the channel number in $i$-th stage.
$N_i$ is the number of building blocks in $i$-th stage.}
\label{tab:architecture}
\small
\renewcommand{\arraystretch}{1.125}
\setlength{\tabcolsep}{2pt}
\begin{tabular}{@{}ccccc@{}}
\toprule
Stage & \makecell{Output\\size} & DTFormer-S & DTFormer-B & DTFormer-L \\
\midrule
1 & $ H/4 \times W/4 $ & \makecell{$C_1=64$\\$N_1=3$} & \makecell{$C_1=80$\\$N_1=4$} & \makecell{$C_1=112$\\$N_1=4$} \\
\addlinespace[3pt]
2 & $ H/8 \times W/8 $ & \makecell{$C_2=128$\\$N_2=4$} & \makecell{$C_2=160$\\$N_2=8$} & \makecell{$C_2=224$\\$N_2=8$} \\
\addlinespace[3pt]
3 & $ H/16 \times W/16 $ & \makecell{$C_3=256$\\$N_3=18$} & \makecell{$C_3=320$\\$N_3=25$} & \makecell{$C_3=448$\\$N_3=25$} \\
\addlinespace[3pt]
4 & $ H/32 \times W/32 $ & \makecell{$C_4=512$\\$N_4=4$} & \makecell{$C_4=512$\\$N_4=8$} & \makecell{$C_4=640$\\$N_4=8$} \\
\midrule
\multicolumn{2}{c}{Decoder dimension:} & 512 & 512 & 1024 \\
\bottomrule
\end{tabular}

\end{table}

\paragraph*{Model architecture}
Tab.~\ref{tab:architecture} provides detailed architectural configurations of DTFormer variants.
$C_i$ represents the channel number in $i$-th stage. $N_i$ is the number of building blocks in $i$-th stage.

\begin{table}[!t]
\centering
\caption{TSA-E parameter sharing configuration. $N_i$ is the total number of blocks in that stage.}
\label{tab:tsam_sharing}
\small
\renewcommand{\arraystretch}{1.125}
\setlength{\tabcolsep}{1pt}
\begin{tabular}{@{}lcccccc@{}}
\toprule
& \multicolumn{2}{c}{DTFormer-S} & \multicolumn{2}{c}{DTFormer-B} & \multicolumn{2}{c}{DTFormer-L} \\
\cmidrule(lr){2-3}\cmidrule(lr){4-5}\cmidrule(lr){6-7}
Stage & \makecell{Share\\Factor} & \makecell{TSA-E\\Instances} & \makecell{Share\\Factor} & \makecell{TSA-E\\Instances} & \makecell{Share\\Factor} & \makecell{TSA-E\\Instances} \\
\midrule
Stage 2 & 2 & \makecell{2 (use\\$N_2=4$)} & 2 & \makecell{4 (use\\$N_2=8$)} & 2 & \makecell{4 (use\\$N_2=8$)} \\
\addlinespace[3pt]
Stage 3 & 4 & \makecell{5 (use\\$N_3=18$)} & 4 & \makecell{7 (use\\$N_3=25$)} & 8 & \makecell{4 (use\\$N_3=25$)} \\
\addlinespace[3pt]
Stage 4 & 2 & \makecell{2 (use\\$N_4=4$)} & 2 & \makecell{4 (use\\$N_4=8$)} & 4 & \makecell{2 (use\\$N_4=8$)} \\
\bottomrule
\end{tabular}

\end{table}

\paragraph*{TSA-E parameter sharing mechanism}
Tab.~\ref{tab:tsam_sharing} illustrates the parameter sharing mechanism of the Text-guided Semantic Alignment for Encoder (TSA-E). We apply a sharing strategy where a group of blocks shares one set of TSA-E parameters. The number of TSA-E instances is calculated as $\lceil N_i / \text{ShareFactor} \rceil$. The decoder (TSA-D) modules remain independent (no sharing) at all levels.

\subsection{Comparison with State-of-the-Art}
\label{sec:comparison}
Tab.~\ref{tab:main_results} compares DTFormer with state-of-the-art methods on NYU Depth V2 and SUN-RGBD. DTFormer consistently outperforms prior methods across all model scales. On NYU Depth V2, DTFormer-S achieves 57.8\% mIoU, improving over DFormerv2-S by 1.8 points with 40.2G FLOPs, while DTFormer-L reaches a new state-of-the-art result of 58.6\%. On SUN-RGBD, DTFormer-S obtains 53.5\% mIoU, a 2.0-point gain over DFormerv2-S, and DTFormer-L further achieves 53.9\%. These consistent improvements across both benchmarks validate the effectiveness of hierarchical text-guided semantic alignment. Training details and qualitative results are provided in Secs.~\ref{sec:experimental_setup} and~\ref{sec:qualitative_results}, respectively.

\begin{figure}[!t]
  \centering
  \includegraphics[width=\linewidth,trim=160 70 220 30,clip]{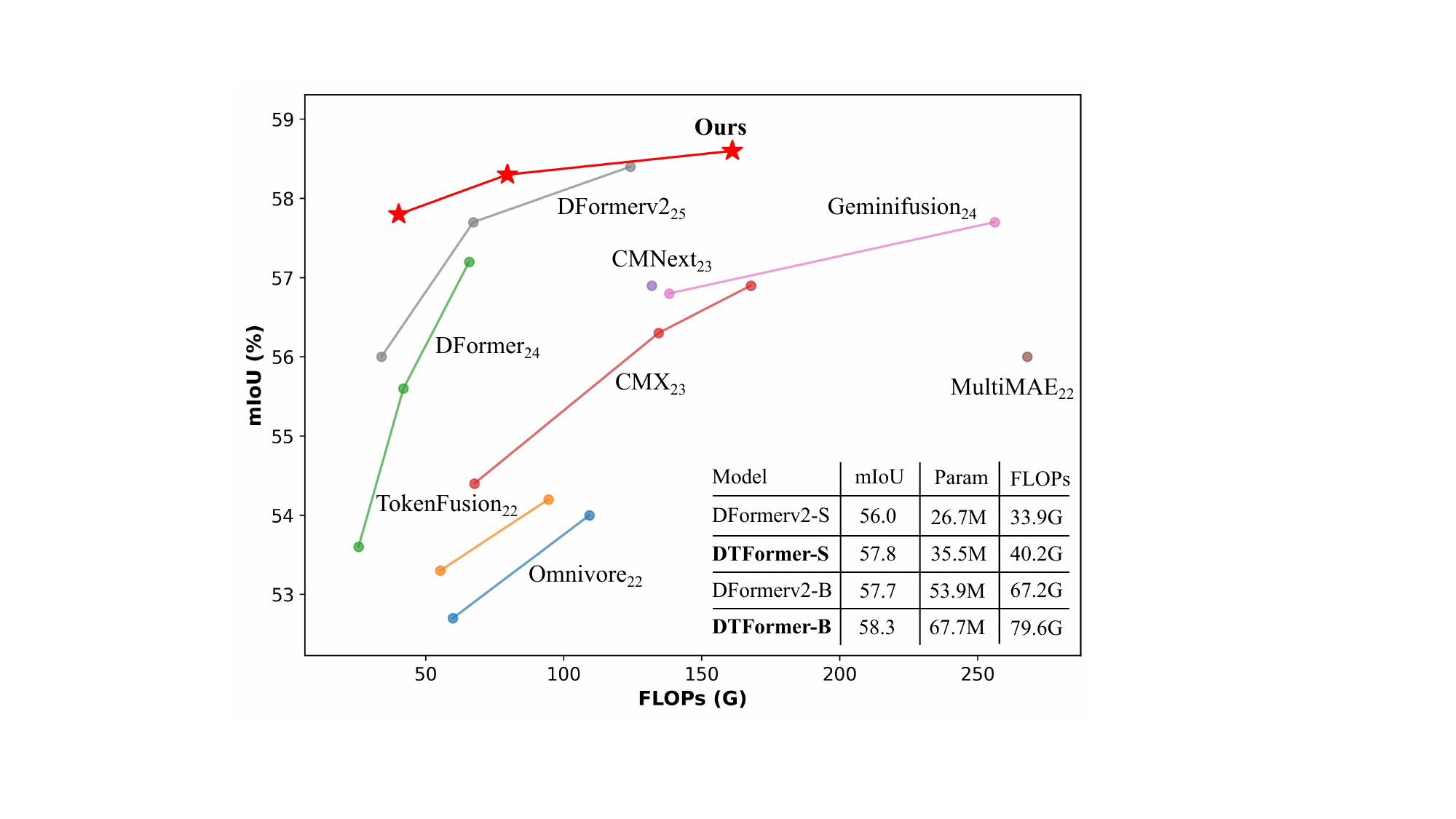}
  \caption{Performance vs.\ FLOPs on NYU Depth V2. DTFormer variants (red) achieve superior mIoU with competitive computational cost.}
  \label{fig:performance_flops}
\end{figure}

\paragraph*{Efficiency analysis}
As illustrated in Fig.~\ref{fig:performance_flops}, DTFormer variants achieve a favorable accuracy–efficiency trade-off compared with dual-encoder methods~\cite{zhang2023cmx,jia2024geminifusion,zhang2023cmnext}. Since the class vocabulary is fixed, the text embeddings of all canonical classes can be precomputed once and stored as a prototype bank, from which scene-relevant prototypes are selected. In temporally continuous and semantically stable scenes, these prototypes may be reused across adjacent frames and refreshed only when the scene context changes, avoiding repeated frame-level prior generation.

\subsection{Ablation Studies}
\begin{table}[!t]
\centering
\caption{Variant composition. Ticks indicate whether encoder TSA-E / decoder TSA-D is enabled.}
\label{tab:ablation_variants}
\small
\renewcommand{\arraystretch}{1.125}
\begin{tabular}{lccc}
\toprule
\textbf{Configuration} & \textbf{TSA-E} & \textbf{TSA-D} & \textbf{mIoU (\%)} \\
\midrule
Baseline & -- & -- & 56.0 \\
+ Decoder only & -- & \checkmark & 56.9~{\footnotesize(+0.9)} \\
+ Encoder only & \checkmark & -- & 57.1~{\footnotesize(+1.1)} \\
\rowcolor{gray!15}
\textbf{DTFormer-S (Full)} & \checkmark & \checkmark & \textbf{57.8}~{\footnotesize(+1.8)} \\
\bottomrule
\end{tabular}

\end{table}

\paragraph*{Effectiveness of TSAM Variants}
Tab.~\ref{tab:ablation_variants} evaluates encoder- and decoder-side text guidance. Starting from the DFormerv2-S baseline of 56.0\% mIoU, TSA-D and TSA-E individually improve performance to 56.9\% (+0.9) and 57.1\% (+1.1), respectively. Combining both variants yields the best result of 57.8\% (+1.8), showing that encoder- and decoder-side text guidance provide complementary benefits. TSA-E supports semantic-aware feature extraction in the encoder, while TSA-D further refines the fused representations during decoding.

\begin{table}[!htbp]
\centering
\caption{Converged gating scalars $\beta_e$ (TSA-E) and $\beta_d$ (TSA-D); ranges/means aggregate across all blocks.}
\label{tab:ablation_beta}
\small
\renewcommand{\arraystretch}{1.125}
\begin{tabular}{lccc}
\toprule
\textbf{Parameter} & \textbf{Init} & \textbf{Converged range} & \textbf{Mean} \\
\midrule
$\beta_e$ (encoder) & 0.50 & 0.54 -- 0.59 & 0.57 \\
$\beta_d$ (decoder) & 0.10 & 0.25 -- 0.27 & 0.26 \\
\bottomrule
\end{tabular}

\end{table}

\paragraph*{Learned injection strength}
Tab.~\ref{tab:ablation_beta} reports the converged values of the per-block gating scalars $\beta_e$ and $\beta_d$. Encoder gates settle at 0.54--0.59, while decoder gates increase from 0.10 to 0.25--0.27, with no block collapsing to a near-zero value. This indicates that both TSA-E and TSA-D actively use text guidance after training, consistent with the ablation in Tab.~\ref{tab:ablation_variants}.

\begin{table}[!t]
\centering
\caption{Encoder injection stages for TSA-E. Ticks indicate which stages receive text guidance.}
\label{tab:ablation_stages}
\small
\renewcommand{\arraystretch}{1.125}
\begin{tabular}{lccccc}
\toprule
\textbf{Configuration} & \textbf{S1} & \textbf{S2} & \textbf{S3} & \textbf{S4} & \textbf{mIoU (\%)} \\
\midrule
Baseline & -- & -- & -- & -- & 56.0 \\
S3--4 only & -- & -- & \checkmark & \checkmark & 56.5~{\footnotesize(+0.5)} \\
S1--4 (all) & \checkmark & \checkmark & \checkmark & \checkmark & 56.6~{\footnotesize(+0.6)} \\
\rowcolor{gray!15}
\textbf{S2--4 (ours)} & -- & \checkmark & \checkmark & \checkmark & \textbf{57.1}~{\footnotesize(+1.1)} \\
\bottomrule
\end{tabular}

\end{table}

\paragraph*{Optimal Injection Stages}
Tab.~\ref{tab:ablation_stages} examines where TSA-E should be injected. Injecting text guidance at stages 2--4 achieves the best result of 57.1\% mIoU, compared with 56.5\% for stages 3--4 and 56.6\% for stages 1--4. This suggests that high-resolution, low-level features at stage 1 are less suitable for category-level language alignment. We therefore apply TSA-E to stages 2--4.

\begin{table}[!htbp]
\centering
\caption{Text guidance strategies. ``Avg.\ Classes'' is the mean number of class names per image.}
\label{tab:ablation_text}
\small
\renewcommand{\arraystretch}{1.125}
\begin{tabular}{lcc}
\toprule
\textbf{Text Strategy} & \textbf{Avg. Classes} & \textbf{mIoU (\%)} \\
\midrule
Fixed class names (all 40) & 40 & 56.1~{\footnotesize(+0.1)} \\
VLM class names ($K=1$) & $\sim$7 & 57.1~{\footnotesize(+1.1)} \\
\rowcolor{gray!15}
\textbf{VLM ($K=2$, ours)} & $\sim$6 & \textbf{57.8}~{\footnotesize(+1.8)} \\
\bottomrule
\end{tabular}

\end{table}

\paragraph*{Text Guidance Strategy}
Tab.~\ref{tab:ablation_text} compares different text priors. Using all 40 fixed class names gives only a marginal gain, 56.1\% mIoU (+0.1), because irrelevant tokens dilute attention. Image-specific VLM predictions are more effective: a single VLM ($K=1$) reaches 57.1\% (+1.1), while our dual-VLM intersection strategy ($K=2$) further improves to 57.8\% (+1.8) with about six class names per image. The intersection reduces hallucinated or model-specific biases while preserving enough semantic coverage; using more than two VLMs ($K>2$) gave diminishing returns because overly strict filtering made the class set too sparse. Our $K=2$ configuration balances precision and recall, validating that image-specific priors are essential for effective text guidance.

\begin{table}[!t]
\centering
\caption{Cross-modal attention design. ``Cosine'': L2-normalized $\mathbf{Q},\mathbf{K}$; ``Temp.'': learnable temperature $\tau$.}
\label{tab:ablation_attention}
\small
\renewcommand{\arraystretch}{1.125}
\begin{tabular}{lccc}
\toprule
\textbf{Configuration} & \textbf{Cosine} & \textbf{Temp.} & \textbf{mIoU (\%)} \\
\midrule
Standard (baseline) & -- & -- & 57.3 \\
+ Cosine sim. & \checkmark & -- & 57.5~{\footnotesize(+0.2)} \\
+ Learnable $\tau$ & -- & \checkmark & 57.4~{\footnotesize(+0.1)} \\
\rowcolor{gray!15}
\textbf{Full (ours)} & \checkmark & \checkmark & \textbf{57.8}~{\footnotesize(+0.5)} \\
\bottomrule
\end{tabular}

\end{table}

\paragraph*{Attention Mechanism Design}
Tab.~\ref{tab:ablation_attention} examines key choices in cross-modal attention. Standard scaled dot-product attention reaches 57.3\% mIoU. Adding cosine similarity or learnable temperature improves performance to 57.5\% (+0.2) and 57.4\% (+0.1), respectively. Combining both achieves the best result, 57.8\% (+0.5), showing that the two designs are complementary: cosine similarity makes attention depend on directional alignment rather than vector magnitudes, while the learnable temperature adapts attention sharpness for more stable and selective text--visual alignment.

\subsection{Analysis of Text Prior Quality}
\begin{table}[!t]
\centering
\caption{Robustness of the oracle-trained DTFormer-S on NYU Depth V2 under progressive text-prior corruption. $\pm$ values denote std over 3 noise realizations.}
\label{tab:ablation_corruption}
\small
\renewcommand{\arraystretch}{1.125}
\begin{tabular}{lcc}
\toprule
\textbf{Setting} & \textbf{mIoU (\%)} & \textbf{$\Delta$ vs.\ baseline} \\
\midrule
\multicolumn{3}{l}{\textit{Reference}} \\
DFormerv2-S & 56.0 & --- \\
\rowcolor{gray!15}
DTFormer-S (ours) & \textbf{57.8} & +1.8 \\
\midrule
\multicolumn{3}{l}{\textit{From oracle, with progressive noise}} \\
Oracle (GT) & 61.8 & +5.8 \\
Replace 1/6 & 59.9~{\footnotesize$\pm0.2$} & +3.9 \\
Replace 2/6 & 57.9~{\footnotesize$\pm0.3$} & +1.9 \\
Replace 3/6 & 56.0~{\footnotesize$\pm0.3$} & 0.0 \\
\bottomrule
\end{tabular}

\end{table}

\label{sec:text_prior_analysis}
\paragraph*{Upper bound with oracle text priors}
To establish the headroom available to the TSAM framework, we replace VLM-generated class names with ground-truth (GT) labels during both training and inference. For a fair cardinality match, the GT prior uses the top six ground-truth categories per image ranked by pixel area, which is close to the average size of the dual-VLM intersection. As shown by the Oracle (GT) row in Tab.~\ref{tab:ablation_corruption}, this oracle setting reaches \textbf{61.8\%} mIoU on NYU Depth V2, a $+5.8$ gain over the DFormerv2-S baseline (56.0\%). The gap between the oracle (61.8) and standard DTFormer-S (57.8) indicates the potential gains from cleaner text priors, and suggests that TSAM can translate more accurate semantic priors into segmentation gains. A per-class breakdown in Tab.~\ref{tab:ablation_confusable} further shows that this $+5.8$ oracle gain concentrates on semantically confusable category pairs (e.g., \textit{desk}/\textit{table}).

\begin{table}[!t]
\centering
\caption{Oracle headroom analysis on semantically confusable NYU Depth V2 class pairs. Baseline denotes DFormerv2-S; Oracle denotes DTFormer-S trained with GT class names.}
\label{tab:ablation_confusable}
\small
\setlength{\tabcolsep}{2pt}
\renewcommand{\arraystretch}{1.125}
\begin{tabular}{@{}llccc@{}}
\toprule
\textbf{Pair} & \textbf{Class} & \textbf{Base.} & \textbf{Oracle} & \textbf{$\Delta$} \\
\midrule
\multirow{2}{*}{window / blinds} & window & 49.8 & 72.2 & +22.4 \\
 & blinds & 61.5 & 83.2 & +21.7 \\
\addlinespace[3pt]
\multirow{2}{*}{desk / table} & desk & 31.4 & 51.9 & +20.5 \\
 & table & 56.2 & 66.4 & +10.2 \\
\addlinespace[3pt]
\multirow{2}{*}{shelves / bookshelf} & shelves & 23.4 & 41.6 & +18.2 \\
 & bookshelf & 49.1 & 61.9 & +12.8 \\
\addlinespace[3pt]
\multirow{2}{*}{dresser / night stand} & dresser & 63.9 & 78.6 & +14.7 \\
 & night stand & 71.2 & 76.1 & \phantom{+}+4.9 \\
\addlinespace[3pt]
\multirow{2}{*}{sofa / chair} & sofa & 65.5 & 74.7 & \phantom{+}+9.2 \\
 & chair & 69.4 & 76.2 & \phantom{+}+6.8 \\
\bottomrule
\end{tabular}

\end{table}

\paragraph*{Per-class analysis on confusable categories}
\label{sec:confusable_categories}
To understand \emph{where} accurate text priors provide the largest headroom, we inspect per-class IoU under the oracle setting (DTFormer-S trained with GT class names, 61.8\%). As shown in Tab.~\ref{tab:ablation_confusable}, the oracle gains over DFormerv2-S are concentrated on semantically confusable pairs---\textit{window}/\textit{blinds}, \textit{desk}/\textit{table}, \textit{shelves}/\textit{bookshelf}, \textit{dresser}/\textit{night stand}, and \textit{sofa}/\textit{chair}. These categories share visual appearance but differ in label, suggesting that accurate class-level priors can complement geometry in resolving semantic ambiguity.

\paragraph*{Robustness to text prior corruption}
Tab.~\ref{tab:ablation_corruption} stress-tests the oracle-trained DTFormer-S by corrupting its text priors only at evaluation time, without retraining. Starting from the oracle GT class names, we progressively replace a fraction of the 6 names per image with random in-vocabulary distractors. Performance degrades smoothly: replacing 1/6 of the names drops mIoU from 61.8 to 59.9 ($\pm$0.2), 2/6 further to 57.9 ($\pm$0.3), and only at 3/6 corruption does performance fall back to the text-free DFormerv2-S baseline (56.0 $\pm$0.3). The graceful degradation suggests that DTFormer retains a degree of robustness to text-prior noise.

\begin{table}[!t]
\centering
\caption{Cross-evaluation under train/test text-source mismatch (GT vs.\ VLM class names).}
\label{tab:ablation_cross_eval}
\small
\renewcommand{\arraystretch}{1.125}
\begin{tabular}{lcc}
\toprule
\textbf{Train~/~Eval} & \textbf{GT text} & \textbf{VLM text} \\
\midrule
GT-trained & 61.8 & 57.5 \\
\rowcolor{gray!15}
VLM-trained (ours) & 58.2 & \textbf{57.8} \\
\bottomrule
\end{tabular}

\end{table}

\paragraph*{Generalization across text sources}
We next evaluate train/test mismatch between GT and VLM class names in all four combinations (Tab.~\ref{tab:ablation_cross_eval}). A GT-trained model evaluated with VLM text retains 57.5\% mIoU, close to the default VLM-trained setting (57.8\%), while a VLM-trained model improves to 58.2\% when given GT text at inference. These results suggest that TSAM is not tightly tied to a particular text source and can benefit from cleaner priors at inference.

\subsection{Qualitative Results}
\label{sec:qualitative_results}

\begin{figure}[!t]
  \centering
  \includegraphics[width=\linewidth,trim=80 70 200 80,clip]{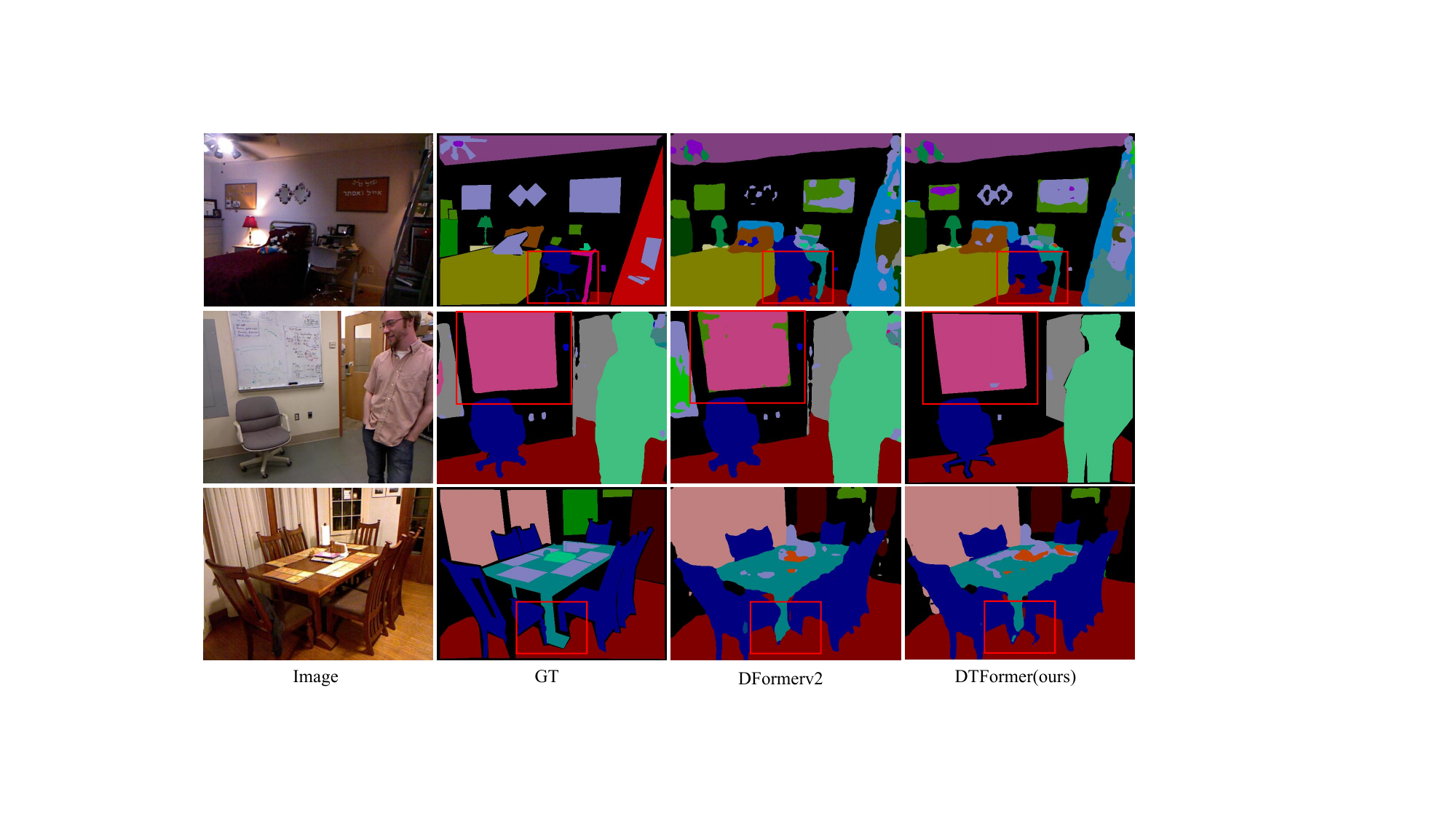}
  \caption{Qualitative comparison on NYU Depth V2.
From left to right: RGB input, ground truth, DFormerv2, DTFormer (ours).}
  \label{fig:qualitative}
\end{figure}

Fig.~\ref{fig:qualitative} presents qualitative comparisons between DTFormer and DFormerv2 baseline on NYU Depth V2. As shown, DTFormer produces improved segmentation across diverse indoor scenes, including bedrooms, offices, and dining rooms. The results demonstrate better handling of challenging structures such as furniture details. These improvements are consistent with the quantitative gains reported in Sec.~\ref{sec:comparison}.

%% file: Sections/5_Conclusion.tex
\section{Conclusion}
\label{sec:conclusion}
We presented DTFormer, a tri-modal framework that brings explicit semantic guidance to RGB-D segmentation through hierarchical text-visual-geometry alignment. Our key insight is that text guidance should be stage-adaptive: lightweight in early encoder stages to preserve geometric reasoning, and enriched in decoder stages for robust semantic-geometric alignment. To address the semantic ambiguity of fixed vocabularies, we leverage vision-language models to generate image-specific class names with intersection-based filtering, ensuring reliable and focused semantic priors. Extensive experiments validate that DTFormer achieves SOTA performance on NYU Depth V2 and SUN-RGBD while maintaining competitive efficiency.